\documentclass[nonacm, sigconf]{acmart}
\AtBeginDocument{%
  \providecommand\BibTeX{{%
    \normalfont B\kern-0.5em{\scshape i\kern-0.25em b}\kern-0.8em\TeX}}}

\usepackage{amsthm}
\usepackage{amsmath}
\usepackage{amsfonts}
\usepackage{hhline}
\usepackage{multirow}
\usepackage{graphicx}
\usepackage{epstopdf}
\usepackage{array}
\usepackage {hyperref}
\DeclareMathAlphabet{\mathcal}{OMS}{cmsy}{m}{n}
\usepackage[mathscr]{eucal}
\DeclareMathAlphabet\mathbfcal{OMS}{cmsy}{b}{n}
\usepackage{multirow}
\usepackage[ruled,vlined,linesnumbered]{algorithm2e}
\usepackage{amsmath}

\newtheorem{definition}{Definition}

\usepackage[font={footnotesize}]{caption}
\usepackage[font={footnotesize}]{subcaption}
\usepackage{cancel}
\usepackage[flushmargin]{footmisc}

\let\oldnl\nl
\newcommand{\nonl}{\renewcommand{\nl}{\let\nl\oldnl}}

\begin{document}
\title{FedAR+: A Federated Learning Approach to Appliance Recognition with Mislabeled Data in Residential Buildings}
\author{Ashish Gupta$^1$, Hari Prabhat Gupta$^2$, and Sajal K. Das$^1$}

\email{{ashish.gupta@mst.edu, hariprabhat.cse@iitbhu.ac.in,  sdas@mst.edu}} 
\affiliation{%
 \institution{$^2$Dept. of Computer Science and Engineering, Indian Institute of Technology (BHU), Varanasi, India}
   }

\affiliation{%
 \institution{$^1$Dept. of Computer Science, Missouri University of Science and Technology, Rolla, MO 65409, USA}
 \vspace{0.5cm}
}

%



\begin{abstract}
With the enhancement of people's living standards and rapid growth of communication technologies, residential environments are becoming smart and well-connected, increasing overall energy consumption substantially. As household appliances are the primary energy consumers, their recognition becomes crucial to avoid unattended usage, thereby conserving energy and making smart environments more sustainable. An appliance recognition model is traditionally trained at a central server (service provider) by collecting electricity consumption data, recorded via smart plugs, from the clients (consumers), causing a privacy breach. Besides that, the data are susceptible to noisy labels that may appear when an appliance gets connected to a non-designated smart plug. While addressing these issues jointly, we propose a novel federated learning approach to appliance recognition, called FedAR+, enabling decentralized model training across clients in a privacy preserving way even with mislabeled training data. FedAR+ introduces an adaptive noise handling method, essentially a joint loss function incorporating weights and label distribution, to empower the appliance recognition model against noisy labels. By deploying smart plugs in an apartment complex, we collect a labeled dataset that, along with two existing datasets, are utilized to evaluate the performance of FedAR+. Experimental results show that our approach can effectively handle up to $30\%$ concentration of noisy labels while outperforming the prior solutions by a large margin on accuracy.

\end{abstract}



\maketitle
\sloppy
\section{Introduction}
Energy consumption in residential buildings is increasing rapidly with the growth of electrical household appliances. According to the United States Energy Information Administration (US EIA)~\cite{EIA}, $22\%$ of the total energy consumption in $2020$ is accounted by residential buildings, requiring dedicated efforts to reduce the usage of electricity. A practical solution is encouraging consumers to use electric appliances efficiently, which involves  recognizing the appliances uniquely based on their consumption patterns recorded via appliance-wise smart plugs~\cite{veloso2019cognitive,soe2019load,reddy2017plug}. 
By having information about currently running appliances, the consumers can minimize the electricity usage by restricting high power appliances (e.g., electric heater, air conditioner) during peak hours~\cite{farrokhifar2018real}. Moreover, the utility company (service provider) may also incentivize the consumers by offering direct monetary benefit through a dynamic pricing policy~\cite{farrokhifar2018real,jang2021offline} and indirect benefit through an appliance-wise breakage of consumption bill. Literature indicates that appliance recognition has been a building block in wide range of important applications such as load forecasting~\cite{demandforecast,wang2021improving}, occupancy detection~\cite{kleiminger2015household}, and energy management in smart buildings~\cite{smartbuild,schwermer2022federated}. However, the current appliance recognition approaches have disregarded the following two practical issues. 

(i) \textit{Privacy preservation of consumers' data} -- As recognition model is essentially a machine learning model, it requires a large amount of labeled training data which, in general, collected from many consumers at a central server (service provider). Sharing of data brings in a privacy concern to the consumers as the data may be misused by adversaries via theft or burglary, and by detecting home occupancy~\cite{allik2020smart, kleiminger2015household}. Hence, the consumers may be reluctant to upload the data and as a consequence, the existing approaches~\cite{yan2019household, vadakattu2018feature, 8580416, soe2019load,veloso2019cognitive, ganu2014socketwatch} would fail to train the recognition model, indicating a need of a model that can be trained collaboratively at the consumer side without sharing any data.

(ii) \textit{Mislabeled training data} -- Some data samples may appear with wrong (noisy) labels when an appliance is mistakenly connected to a non-designated smart plug.\footnote{Assuming that each smart plug, during deployment, is designated to a specific appliance to collect the labeled data automatically.} Moreover, a compromised consumer may also flip the labels in its local dataset with an intent to poison the model. Such consumers might receive monetary benefits from rival service provider. As prior studies~\cite{vadakattu2018feature, 8580416, soe2019load,veloso2019cognitive} do not incorporate any noisy-label handling mechanism, they can not withstand mislabeled training data. Slightly on a different track, learning with noisy labeled data has been a topic of great interest in computer vision; however, the proposed solutions~\cite{s1,s2,s3,s4,s5, gao2017deep} mainly relied upon visual features, thus their applicability to time series data (generated from smart plugs) is discouraged. 

Although there exist some works~\cite{schwermer2022federated,qureshi2022poisoning} on privacy-preserving appliance recognition using Federated Learning (FL), they do not consider the presence of noisy labels in training data. {\em In this paper, we address this important challenge by building an appliance recognition model in a collaborative manner using mislabeled training data while preserving consumers' privacy.} To the best of our knowledge, this is the first work to tackle the practical issues of privacy preservation and mislabeled training data \textbf{jointly} for appliance recognition in residential buildings.

\noindent \textbf{Contributions:} 
Major contributions of this paper are given below:
\begin{itemize}
\item With a goal to train an appliance recognition model across distributed consumers using their local private data, we propose a novel federated learning approach, called FedAR+, in presence of a coordinating server (service provider).  
  The server initializes the training by broadcasting the model (i.e., weights) to all the clients (consumers); each client re-trains the model using its local data and dispatches the updated model back to the server for aggregation. By repeating the above steps for some iterations, FedAR+ produces a generalized model without exposing the consumers' data. 
  \item FedAR+ incorporates an innovative aggregation function to deal with the biasing problem caused due to non independent and identically distributed (non-iid) data across clients. 
 
 \item We propose an adaptive noise handling method that strategically exploits a joint loss function, incorporating the weight parameters and label distributions, to enable the learning with mislabeled training data.
 \item Finally, we collect real data by deploying smart plugs in three houses in an apartment complex, to experimentally validate the performance of FedAR+. Moreover, to demonstrate its efficacy, we also employ two widely used datasets from the same domain. The overall results show that FedAR+ outperforms prior solutions by a large margin while achieving an accuracy of more than $86\%$ even when the concentration of noisy labels in training data is as high as $30\%$.  

\end{itemize}
   
\noindent The paper is organized as follows. Section~\ref{related-work} reviews the related work while Section~\ref{flapproach} proposes our federated learning approach, FedAR+. Section~\ref{prelim} discusses the dataset creation steps and elaborates the causes for the presence of noisy labels. Section~\ref{model} builds the underlying appliance recognition model with a noise handling method. Section~\ref{results} evaluates the performance of FedAR+ and compares with prior solutions. Finally, Section~\ref{conclude} concludes the paper. 

\section{Related Work}\label{related-work}
This section discusses the notable and relevant existing works to position our proposed approach.

\vspace{-0.1in}
\subsection{Appliance recognition} Many works exist on appliance recognition as it has been a building block to energy monitoring applications. For example, in~\cite{smartbuild}, a lightweight appliance recognition model is developed for energy management in smart buildings. The authors in~\cite{ganu2014socketwatch} attempted to identify a malfunctioning appliance and its operating states by leveraging electricity consumption patterns. While a line of works~\cite{yan2019household, veloso2019cognitive,8580416} involved in distinguishing the appliances from one to another, the work in~\cite{soe2019load} aimed to identify load profile as intermittent, continuous, or phantom, for energy management in smart home settings. Slightly different from above works, Codispoti \textit{et al.}~\cite{codispoti2022learning} presented a $K$-active neighbors based appliance recognition approach to learn from unlabeled data collected via Arduino operated smart plugs.   

Although the aforementioned prior approaches achieve good performance using machine learning and deep learning algorithms, their performance heavily relies on the assumption that the training data are correctly labeled and do not contain any noisy labels. However, in practice, satisfying this assumption requires additional care from the consumers during data collection (via smart plugs~\cite{ganu2014socketwatch,yan2019household, veloso2019cognitive,codispoti2022learning}), restricting their flexibility and thereby the consumers may be reluctant to adopt such solutions. Besides that the recognition model should not fully rely upon the consumers' actions rather it should be robust enough to leverage mislabeled training data.

\vspace{-0.1in}
\subsection{Learning with noisy labels} 
Literature indicates that learning with mislabeled (noisy labeled) training data has been a widely studied problem in computer vision and image processing because the manual labeling is time consuming and costly~\cite{song2022learning,liang2022few}. The work in~\cite{s1} presented an iterative learning approach to re-label the noisy-labeled training samples while in another work~\cite{s2} the authors estimated correct labels against noisy ones during training by jointly optimizing the model parameters and intermediary corrected labels.  In~\cite{s4}, a symmetric learning approach is proposed to simultaneously address the presence of noisy labels and overfitting problem of Deep Neural Networks (DNNs). A distillation process leveraging knowledge graph is introduced in~\cite{s3} to learn with noisy labels. Recently, a meta-learning approach is developed in~\cite{s5} to directly learn correct labels from the training data. However, as the prior approaches mostly work around visual features, they can not offer an accurate solution to mislabeled time series data. 

\vspace{-0.1in}
\subsection{Federated learning}
In last few years, a new learning paradigm, Federated Learning (FL)~\cite{mcmahan2017communication} has received an unprecedented attention because it facilitates collaborative model training without compromising clients' privacy. Prior works illustrate the effectiveness of FL in real-world applications such as next word prediction~\cite{hard2018federated}, keyword spotting~\cite{leroy2019federated}, and visual object detection~\cite{liu2020fedvision}. However, FL is yet to be explored for the appliance recognition models that are otherwise trained at the central server by collecting data from multiple clients (consumers) and revealing the client's privacy. FL offers an effective solution to this problem by keeping the data locally with the clients while allowing participate in collaborative training of the model. Recently, a few studies~\cite{qureshi2022poisoning,zhang2022fednilm} have also attempted to apply FL in smart energy management to enable load forecasting and load disaggregation at consumer side. In another work~\cite{schwermer2022federated}, the authors presented an FL approach to identify office plug load, however they do not consider the presence of noisy labels in the training data, which we aim to address in this work. Besides all, the application of FL to appliance recognition needs to be investigated from robustness perspective in the presence of noisy labels.   






\vspace{-0.05in}
\section{F\lowercase{ed}AR+ Approach}\label{flapproach}
This section presents an overall setup of our FL approach, FedAR+, with multiple clients\footnote{A client refers to a low-end computing device (e.g., personal computer) installed at consumer's house to collect data from smart plugs. The device is capable enough to train the underlying appliance recognition model.} and a common remote server, as depicted in Figure~\ref{fl_arch}.  
In appliance recognition scenario, the consumer acts as a client and the service provider works as a remote server. A client may have many appliance-specific smart plugs, each connected to a designated appliance to measure the appliance's electricity consumption and transfer that data to a local in-house computing device. 
To initialize training, the server dispatches an appliance recognition model to all the clients. Each client retrains the model using its local data and sends the weight updates to the server for aggregation. Next, the aggregated (or global) model is sent back to the clients. By repeating the above steps for a certain number of global rounds, the model eventually converges to an optimal solution. With this, FL helps achieve a more generalized and accurate model without sharing the client's local data. 

In FedAR+, we build a deep learning model for appliance recognition which requires a large amount of data for training; however, at the beginning of the deployment, the clients may not have sufficient data. Therefore, we consider the availability of an auxiliary dataset with the clients before the initialization of FL training. To avoid flow disruption, we discuss the auxiliary dataset (collected beforehand from some anonymous houses) in Section~\ref{prelim} and appliance recognition model with noise handing method in Section~\ref{model}, separately. Besides that, we formulate an aggregate function to alleviate the bias that might be introduced by the clients having substantially larger dataset than the others.



\begin{figure}[h]
\vspace{-0.1in}
\centering
\includegraphics[scale=0.9]{Figures/FL.eps}
\vspace{-0.1in}
\caption{Overview of FedAR+. At each client, a noise handling method is incorporated to enable the model learn with mislabeled data.}
\label{fl_arch}
\vspace{-0.1in}
\end{figure}


Let $K$ denote the number of clients collaborating in the learning to build the recognition model. These clients need not be the same from which the auxiliary data ware collected.
Let $\mathcal{D}^j =\{\mathbfcal{X}, \mathbfcal{Y}\}$ be the dataset with $j^{th}$ client, which includes both auxiliary and local data collected over a fixed period of time, where $1 \leq  j \leq K$. At each update round, the objective of the remote server is to learn optimal weight parameters $\pmb{\theta}$ by minimizing an empirical loss function as
\begin{align}
 \underset{\pmb{\theta}}{\text{argmin}} \quad \left \{ \mathcal{F}(\pmb{\theta}) = \Psi \left(\{f^j(\pmb{\theta})\}_{1 \leq j \leq K}\right) \right \}, 
\end{align}
\vspace{-0.1in}

\noindent
where $\Psi(\cdot)$ is an aggregate function and $f^j(\cdot)$ is the local objective function used by the $j^{th}$ client. We also propose a noise handling method, in Section~\ref{adaptive}, to facilitate learning with mislabeled training data at the client.


\vspace{-0.1in}
\subsection{\textbf{Local model update at client}} FedAR+ uses second-order method to perform local updates at the client. Particularly, we adopt canonical Newton's method of the form $-\nabla^2 (f^j)^{-1}\nabla f^j$~\cite{nocedal2006numerical}
as it improves the convergence rate and reduces the accumulation of errors. 
Given the weight parameters $\pmb{\theta}_{[t]}$ of the global model at update round $t$, the client $j$ first computes the local gradient as 

\vspace{-0.15in}
\begin{align}
 g^{j}_{[t]} =\nabla f^j\big( \pmb{\theta}_{[t]} \big).
\end{align}
The client next computes the second-order gradient (Hessian matrix) at $\pmb{\theta}_{[t]}$ as follows

\vspace{-0.15in}
\begin{align}
  h^{j}_{[t]} =\nabla^2 f^j\big( \pmb{\theta}_{[t]} \big).
\end{align}
Now, the local model at the client $j$ is updated as 
 \begin{align}\label{clientupdate}
 \pmb{\theta}_{[t+1]}^j = \pmb{\theta}^{j}_{[t]} - \eta (h^{j}_{[t]})^{-1} g^{j}_{[t]} 
\end{align}
and $\eta$ is the learning rate. Finally, the local updates are sent back to the server for aggregation. 

\vspace{-0.1in}
\subsection{\textbf{Global model update at server}} The problem of aggregation at the server becomes quite simple if we assume that all the clients have independent identically distributed (iid) data, and it can be easily solved by using FedAvg~\cite{mcmahan2017communication} as 
\begin{align}\label{fedavg}
 \Psi \left(\{f^j(\pmb{\theta})\}_{1 \leq j \leq K}\right) \overset{\underset{\mathrm{def}}{}}{=} \sum_{j=1}^{K} \frac{N^j}{N^{total}} \pmb{\theta}_{[t+1]}^j,
\end{align}
where $N^j$ is the size of $\mathcal{D}^j$ and $N^{total}=N^1 + N^2 + \cdots + N^h$. However, this assumption is unrealistic for applying FL to appliance recognition as the clients may have different number of appliances (essentially non-iid data). With FedAvg, the model may be biased towards the clients who have substantially larger dataset than others.  
To deal with this situation, we introduce an aggregation function
\begin{align} \label{agg}
 \Psi \left(\{f^j(\pmb{\theta})\}_{1 \leq j \leq K}\right) \overset{\underset{\mathrm{def}}{}}{=} \sum_{j=1}^{K} \frac{1}{K} \cdot \pmb{\theta}_{[t+1]}^j. 
\end{align}
With the new aggregation function, each client would receive an unbiased model regardless of number of appliances the client possesses. Algorithm~\ref{algorithm1} summarizes the major steps of FedAR+ with $K$ clients for $T$ number of global rounds.

\SetAlFnt{\small}
\begin{algorithm}[h]
\caption{FedAR+}
\label{algorithm1}
\nonl \textbf{Initialization:} \\
The server builds and broadcasts a recognition model along with an auxiliary dataset to all $K$ clients. \\

\For{$t\gets 1$ to $T$}{
\smallskip
\nonl \textbf{Local model update at round $t$:}\\
\For{each client $j \in \{1, 2, \cdots, K\}$}{
\nonl /* $\pmb{\theta}^j_{[t]}$ is the weight parameters of local model */ \\
Obtain $\pmb{\theta}^j_{[t+1]}$ using Eq.~\ref{clientupdate}. // \textit{the underlying loss function is formulated in Section~\ref{adaptive}}. \\
Dispatch $\pmb{\theta}^j_{[t+1]}$ to the server.   
}
\smallskip
\nonl \textbf{Global model update using aggregation at round $t$:}\\
$\pmb{\theta}_{[t+1]} = \sum_{j=1}^{K} \frac{1}{K} \cdot \pmb{\theta}_{[t+1]}^j$, using Eq.~\ref{agg}.\\
Broadcast $\pmb{\theta}_{[t+1]}$ to the clients
}
\end{algorithm}

\noindent $\bullet$ \textit{Model convergence:} The global recognition model (at the server) advances as the training progresses and it is said to be converged when stops advancing or reaches to the optimal solution. We study the model convergence, under the standard assumptions on the function $\mathcal{F}(\cdot)$~\cite{mcmahan2017communication}, in terms of the optimality gap $\delta = \mathcal{F}(\pmb{\theta}_{[T]}) - \mathcal{F}(\pmb{\theta}^*)$, where $T$ denotes the maximum number of rounds and $\pmb{\theta}^*$ denotes the weights of the optimal model. Ideally, $\delta \approx 0$ for a sufficiently large $T$. Assuming $K$ clients, FedAR+ can achieve $O(\frac{1}{\sqrt{KT}})$ convergence for our DNN model (i.e., non-convex optimization problem). Our experimental results, reported in Section~\ref{result1}, show that the global model converges in $T=30$ rounds (with $50$ local iterations on each clients at each round) with $10$ clients even when there exist $30\%$ noisy labels. The theoretical proof of the convergence rate and the theoretical bounds over the local gradients with heterogeneous clients are to be investigated in our future work.


\section{Auxiliary Dataset}\label{prelim}
In FedAR+, the server is assumed to provide an auxiliary dataset to the clients, before initializing the training, to enable them for an effective local update of the model.  We therefore first discuss data collection and preprocessing for creating an auxiliary dataset. We utilize the power consumption data for recognizing appliances such as refrigerator, electric kettle, television, etc. The data are collected by connecting the appliance to power socket through a designated smart plug that provides a sequence of time stamped readings at a preset sampling rate.   

\begin{definition}[\textbf{Time Series of Consumption}]
It is a temporal sequence of data points collected over a period of time. Let $X =\{x_1, x_2,\cdots, x_n\}$ denote the Time Series of power Consumption (TSC) readings from a designated smart plug, where $n$ is the total number of data points collected during the entire experiment; and  $x_i$ denotes a reading taken at time $t_i$, where $1 \leq i \leq n$ and $t_{i-1} < t_i$. 
\end{definition}

\vspace{-0.15in}
\subsection{Dataset creation}\label{data_prep}
We construct a dataset using TSCs of different appliances. As the appliance can change its state from ON to OFF or vise-versa several times, each TSC (denoted by $X$) includes readings corresponding to both the states. We first separate out only the subsequences (of $X$) that correspond to ON states occurred at distinct time steps along $X$. Then for each separated subsequence, an appliance footprint is computed and stored as an instance of the respective appliance. 

\subsubsection{\textbf{Data collection}} To create an auxiliary dataset, we collected power consumption data from three different houses (within an apartment complex) where each possesses six common household appliances: refrigerator, microwave oven, television, washing machine, air conditioner, and mixer grinder. Each appliance is connected to a designated smart plug that transmits the readings to a in-house data collector (e.g., personal computer) at 1Hz. As we collected the data for a period of one month from each house, we got total $18$ TSCs (i.e., six time series from each of the three houses).    

\subsubsection{\textbf{Preprocessing}}

Let us first discuss the terminology for better illustration of data preprocessing.\\ 
$\bullet$ \textit{Switch point:}
For a TSC $X=\{x_1, x_2, \cdots, x_n\}$, a time instance $t$ is said to be a switch point if the following conditions hold: 
(i) The difference $\delta(t) = |X(t) - X(t-1)| > \phi_1$, a predefined threshold, where $X(t)$ and $X(t-1)$ denote the power consumption readings at time $t$ and $t-1$, respectively; and 
(ii) The rate of change in power readings $\delta_r(t) = \delta(t) / X(t) > \phi_2$, another threshold. 
For setting an appropriate value for $\phi_1$ and $\phi_2$, we visualized several time series for different appliances including both low power (e.g., television) and high power (e.g., air conditioner). We observed that with $\phi_1=30$ watts and $\phi_2=0.2$ (i.e., $20\%$) jointly, the switch points can be detected correctly for most commonly available appliances such as refrigerator, microwave oven, air conditioner, etc. Further, as the thresholds are set empirically, their values are subject to change according to the appliance' operating environment (such as brand and power rating standards of different countries). With small thresholds, we may get frequent false positive; on the contrary, some ON states may get lost with large thresholds. 
 


\noindent $\bullet$ \textit{Steady point:} A time point $t$ along the time series $X$ is said to be steady if $\delta_r(t) < \phi_2$. 

\noindent $\bullet$ \textit{Steady period:} Given a time series $X$, a steady period is a subsequence $X_{t:m}=\{x_{t+1}, x_{t+2}, \cdots, x_{t+m}\}$ if all of its time points are steady. Here, $t$ and $m$ respectively denote a switch point and the length of the steady period, where $t < (n-m)$. 

\begin{definition}[\textbf{Appliance Footprint}]
For a given steady period $X_{t:m}$, corresponding to the ON state of the appliance, we define the appliance footprint as: 
\begin{align}
 X_{af} = \{ X_{t:m}(i) - X_{t:m}(i-1) \ | \ 1 \leq i \leq m\},
\end{align}
where $X_{t:m}(i)$ denotes the $i^{th}$ data point of steady period.
\end{definition}

We compute single-order differences between the consecutive data points to capture subtle fluctuations, revealing better identifiable patterns than those with higher-order statistics. Moreover, the single order difference automatically scales down the values to a smaller range, eliminating the need of normalization. The appliance footprint essentially represents the power consumption pattern of the appliance when it is active. 
For a given TSC $X=\{x_1, x_2, \cdots, x_n\}$, the extraction of footprints includes following three steps:
\begin{enumerate}
 \item Identify switch points in $X$ under thresholds $\phi_1$ and $\phi_2$. 
 \item For each identified switch point $t$, follow two sub-steps:
  (a) search for a steady period of length $m$ after $t$. Let $X_{t:m} = \{x_{t+1}, x_{t+2}, \cdots, x_{t+m}\}$ be a steady period obtained after the switch point $t$. 
 (b) if $X(t) - X(t+m) < 0$, then the steady period $X_{t:m}$ corresponds to ON state of the appliance; otherwise OFF state.
 \item The steady periods corresponding to ON states, are used to obtain appliance footprints; each of which along with its label (name of the appliance) is stored as an instance. 
\end{enumerate}

\begin{figure}[h]
\centering
\includegraphics[scale=0.95]{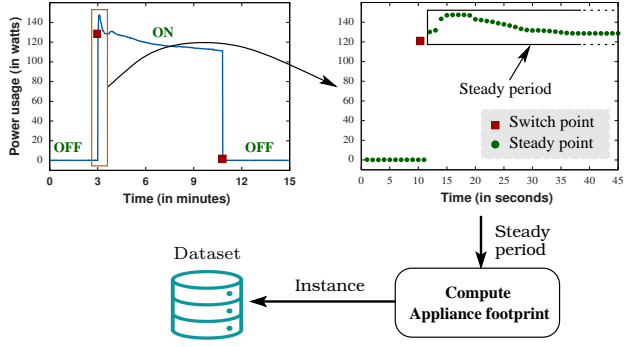}
\vspace{-0.1in}
\caption{Illustrating switch points, steady points, and steady period, in a TSC of a refrigerator for a window of $15$ minutes. }
\label{state}
\vspace{-0.1in}
\end{figure}

\noindent Figure~\ref{state} illustrates a TSC of a refrigerator with identified switch points, steady points, and steady period. The obtained steady periods are used to compute the appliance footprints. Upon obtaining the footprints by processing TSCs of all the appliances, we perform padding on shorter instances to make all the instances of equal length and store them in the dataset. which later provided to the clients as an auxiliary dataset.

\vspace{-0.1in}
\subsection{Presence of noisy labels}
During local data collection at the client, some TSCs may get associated with noisy (wrong) labels due to following reasons:
\begin{itemize}
     \item {\em From deployment perspective:} An appliance (say $A_1$) mistakenly got connected to a non-designated smart plug that was marked to connect with some other appliance (say $A_2$). Consequently, the generated TSC receives a noisy label $A_2$, creating several mislabeled instances (wrong appliance footprints) in the training dataset. It is true that such noisy labels may be avoided {\em at the cost of additional care from the consumers}, however it is not preferable rather the model should be robust against mislabeled training data.
     \item {\em From security perspective:} A malicious consumer (or compromised client) may attempt to inject a wrong label intentionally, to gain some incentive from a rival service provider. As a consequence, the client's local model would generate corrupted local updates, which eventually would diminish the performance of global model.
\end{itemize}


\begin{definition}[\textbf{Noisy label}] Let $\mathcal{D} = \{(\mathbfcal{X}, \mathbfcal{Y})\}$ be a dataset where $\mathcal{X}^i \in \mathbfcal{X}$ is the $i^{th}$ instance associated with a class label $\mathcal{Y}^i \in \mathbf{Y} = \{y_1,y_2,\cdots, y_C\}$, the set of all $C$ classes (appliances). The label $\mathcal{Y}^i$ is said to be noisy if either of the following holds: (i) $\mathcal{Y}^i$ is mislabeled as other class label, i.e., $\mathcal{Y}^i \in \{\mathbf{Y}- y_c\}$, where $y_c$ denotes the correct class label of $\mathcal{X}^i$, or (ii) $\mathcal{Y}^i$ is an arbitrary class label, i.e., $\mathcal{Y}^i \notin \mathbf{Y}$.

%
\end{definition}


\section{Appliance Recognition Model}\label{model}
This section presents a deep neural network (DNN) for appliance recognition that trains collaboratively on the locally collected data and the auxiliary dataset provided by the server. 
The choice of DNN is inspired by its success at recognition tasks with a rich set of learnable features. The network (or model) learns from a training dataset and predicts the class label of a new instance. On the top of that, we propose an adaptive noise handling method to enable the model learning with mislabeled training data at the clients.  



\vspace{-0.1in}
\subsection{Base model overview}
\begin{figure}[h]
\vspace{-0.1in}
\centering
\includegraphics[scale=0.95]{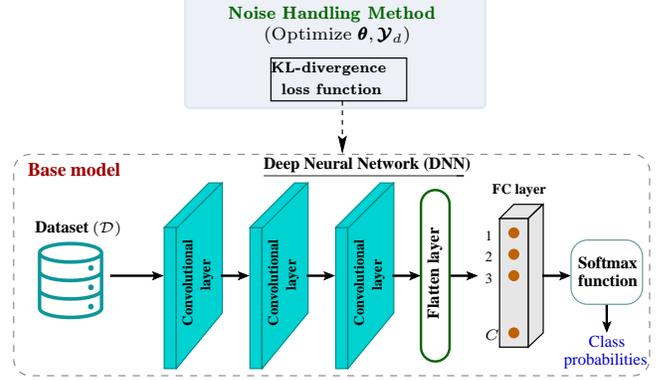}
\vspace{-0.2in}
\caption{An overview of the appliance recognition model in FedAR+.}
\label{dnn}
\vspace{-0.3cm}
\end{figure}

We build a DNN with three convolutional layers (connected sequentially) followed by a flatten and a Fully Connected (FC) layer, as shown in Figure~\ref{dnn}. Let $\mathcal{D}=\{\mathbfcal{X}, \mathbfcal{Y}\}$ be a training dataset (including auxiliary dataset) with the client. The model takes a training dataset $\mathcal{D}$ and yields a set of class probabilities using a \textit{softmax} function. The convolutional layers are all one-dimensional, each consisting of $128$ filters of size $1\times 1$ with input shape $(1, m)$, where $m$ denotes the number of data points in each instance (i.e., appliance footprint). 
Considering there exist total $C$ labels in $\mathcal{D}$, we use $C$ neurons at the FC layer. Finally, a \textit{softmax} function is applied on the output of FC layer to get the class probabilities.          


Now, we present mathematical formulation of our base model (i.e., excluding the noise handling method).  
Given the dataset $\mathcal{D}$, the recognition model mainly attempts to learn a mapping $\mathcal{H}: \mathbfcal{X} \rightarrow \mathbfcal{Y}$, which usually expressed as  
\begin{align}\label{mapping}
 \mathcal{H}(\mathbfcal{X}, \pmb{\theta})= \sigma_{\mathbf{Y}} (\pmb{\theta} \mathbfcal{X}), 
\end{align}
where $\sigma(\cdot)$ is a \textit{softmax} function and $\mathbf{Y}=\{y_1, y_2, \cdots, y_C\}$ is a set of different class labels in $\mathcal{D}$. The function $\sigma(\cdot)$ can transform a vector into probability distribution over its elements. For a vector $z \in \mathbb{R}^C$, the \textit{softmax} function is:
\begin{align}\nonumber
 \sigma_c (z) &= \frac{\text{e}^{z_c}}{\sum_{c=1}^{C}\text{e}^{z_c}} = p(c|z) \quad \forall c \in \{1,2,\cdots, C\}.
\end{align}
Rewriting Eq.~\ref{mapping},

\vspace{-0.1in}
\begin{align}\label{mapping1}
 \mathcal{H}(\mathbfcal{X}, \pmb{\theta}) = p(\mathbf{Y} | \mathbfcal{X}, \pmb{\theta}).
\end{align}
\vspace{-0.1in}

To this end, the appliance recognition problem with base model, at the client, can be observed as a local optimization problem

\vspace{-0.1in}
\begin{align}\label{opt}
 f(\pmb{\theta}) = \underset{\pmb{\theta}}{\text{argmin }} \big \{\mathcal{L} \big (\mathbfcal{Y}, \mathcal{H}(\mathbfcal{X}, \pmb{\theta}) \big) \big\},
\end{align}
\vspace{-0.1in}

\noindent
where $\mathcal{L}(\cdot)$ is an underlying empirical loss function. The base model employs cross-entropy loss function as widely used in DNNs for solving recognition problems~\cite{goodfellow2016deep}. The cross-entropy loss function for $\mathcal{D}$ with $N$ instances, is given as

\begin{small}
\vspace{-0.1in}
\begin{align}\label{loss} \nonumber
 \mathcal{L} \big (\mathbfcal{Y}, \mathcal{H}(\mathbfcal{X}, \pmb{\theta}) \big) &= - \frac{1}{N} \sum_{i=1}^{N} \sum_{c=1}^{C} \mathbb{I}(\mathcal{Y}^i, y_c)\text{log} \ \mathcal{H}\big (\mathcal{X}^i, \pmb{\theta} \big ), \\
 & = - \frac{1}{N} \sum_{i=1}^{N} \sum_{c=1}^{C} \mathbb{I}(\mathcal{Y}^i, y_c) \text{log} \ p\big(\mathcal{Y}^i =y_c \big | \mathcal{X}^i, \pmb{\theta}\big),
\end{align} 
\vspace{-0.1in}
\end{small}

\noindent where $\mathbb{I}(\mathcal{Y}^i, y_c) = 1$ if $\mathcal{Y}^i$ is $y_c$, and $0$ otherwise.
\subsection{Noise handling method}\label{adaptive}
To enable the base model learning from mislabeled training data, this section proposes a noise handling method that       
learns correct labels by iteratively updating the label distributions probabilistically, which is significantly different from the existing approaches~\cite{xu2022trusted,li2022unimodal,gao2017deep} where constant distributions were used in all the iterations. In our method, the model optimizes label distributions along with weight parameters during training, and therefore the local objective function (i.e., Eq.~\ref{opt}) can be written as  
 \begin{align}\label{adaptobj}
 f(\pmb{\theta}) = \underset{\pmb{\theta}, \mathbfcal{Y}_d}{\text{argmin }} \big \{\mathcal{L} \big (\mathbfcal{Y}_d, \mathcal{H}(\mathbfcal{X}, \pmb{\theta}) \big) \big\},
\end{align}

\noindent where $\mathbfcal{Y}_d$ denotes the label distributions among $C$ classes for all $N$ instances of the dataset $\mathcal{D}$. To solve Eq.~\ref{adaptobj}, we introduce a noise handling method consisting of three steps explained below. 

 \subsubsection{Learn weight parameters $\pmb{\theta}$}
First, we train the base model with cross-entropy loss function (Eq.~\ref{loss}) on the training dataset $\mathcal{D}$. By optimizing the loss function, the model learns the weight parameters $\pmb{\theta}$. Due to noisy labels, the learned weights may be far from optimality; nevertheless, they can certainly be used for the initial estimation of the label distributions over the training data. 

 \subsubsection{Estimate label distributions $\mathbfcal{Y}_d$:}
Given the trained model, the label distributions $\mathbfcal{Y}_d$ can be estimated for all instances of $\mathcal{D}$ through validation as $ \mathbfcal{Y}_d = \mathcal{H}(\mathbfcal{X}, \pmb{\theta})$. For each instance, the model provides probability distribution of labels using learned $\pmb{\theta}$. The label with highest probability is assigned to the instance. In general, if the assigned label is the same as true, then its probability should differ substantially from that of other labels. However, this statement holds only if the training dataset does not contain any noisy label. Hence, our method utilizes the distribution instead of only the highest probable label while computing the loss. 

 \subsubsection{Optimize $\pmb{\theta}$ and $\mathbfcal{Y}_d$:}
This step aims to optimize $\mathbfcal{Y}_d$ using Kullback-Leibler (KL) divergence~\cite{mackay2003information} and subsequently fine-tuning the parameters $\pmb{\theta}$ with optimized version of $\mathbfcal{Y}_d$. The KL-divergence measures the difference between two probability distributions. Thus, the loss function of the base model, defined Eq.~\ref{loss}, is replaced by

\vspace{-0.2in}
\begin{align}
 \mathcal{L} \big (\mathbfcal{Y}_d, \mathcal{H}(\mathbfcal{X}, \pmb{\theta}) \big) = \frac{1}{N}  \sum_{i=1}^{N} & KL (\mathcal{Y}_d^i \parallel \mathcal{H}(\mathcal{X}^i, \pmb{\theta}) ),\\
 \text{where} \quad KL (\mathcal{Y}_d^i \parallel \mathcal{H}(\mathcal{X}^i, \pmb{\theta}) ) =& \sum_{c=1}^{C} \mathcal{Y}_d^{i,c} \ \text{log} \left( \frac{\mathcal{Y}_d^{i,c}}{\mathcal{H}_c(\mathcal{X}^i, \pmb{\theta})}\right). \nonumber
\end{align}
Let us first compute the gradient of $\mathcal{L}(\cdot)$ for all $i$ and $c$ as
\begin{align}
 \frac{d\mathcal{L} \big (\mathbfcal{Y}_d, \mathcal{H}(\mathbfcal{X}, \pmb{\theta}) \big)}{d\mathcal{Y}_d^{i,c}} = 1 + \sum_{c=1}^{C} \ \text{log} \left( \frac{\mathcal{Y}_d^{i,c}}{\mathcal{H}_c(\mathcal{X}^i, \pmb{\theta})}\right),
\end{align}
and then update $\mathbfcal{Y}_d$ by
\begin{align}
 \mathbfcal{Y}_d = \mathbfcal{Y}_d - \eta \frac{d\mathcal{L} \big (\mathbfcal{Y}_d, \mathcal{H}(\mathbfcal{X}, \pmb{\theta}) \big)}{d\mathbfcal{Y}_d},
\end{align}
where $\eta$ is the learning rate. Once $\mathbfcal{Y}_d$ stabilizes, the model stops learning. With learned $\mathbfcal{Y}_d$, the weight parameters $\pmb{\theta}$ are then fine tuned for a fixed number of local iterations usually preset by the server before initializing the training. Finally, the updated weights $\pmb{\theta}$ are collected by the server from all the clients for aggregation. Note that the proposed noise handling method is {\em adaptive} as it automatically adapts to mislabeled training data without requiring any additional mechanism to correct the labels beforehand.

\vspace{-0.05in}
\subsection{Prediction}
\begin{figure}[h]
\vspace{-0.1in}
\centering
\includegraphics[scale=0.97]{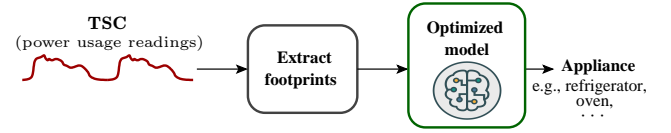}
\vspace{-0.2in}
\caption{Appliance recognition using the optimized recognition model $\pmb{\theta}^*$.}
\label{predict}
\vspace{-0.1in}
\end{figure}

Given the trained model and a testing TSC (generated from the smart-plug), we first need to extract appliance footprints using the preprocessing steps (discussed in Section~\ref{data_prep}) to prepare input to the model.
Let $\pmb{\theta}^*$ be the optimized model, obtained after $T$ global rounds, dispatched from the server to all the clients to predict the class label (or recognitze an appliance) using the footprint.  Figure~\ref{predict} illustrates the recognition process using DNN-based model.       

Let $\mathcal{X}{'} \notin \mathcal{D}$ be a testing instance (an appliance footprint) for which the class label is to be predicted. The client utilizes the model with $\pmb{\theta}^*$ to first compute the posterior class probabilities and then assigns the highest probable class label to $\mathcal{X}{'}$ as expressed below 
\begin{align}
\vspace{-0.1in}
 \mathcal{Y}{'} = \underset{y_c}{\text{argmax}} \ \{\sigma_{y_c} (\pmb{\theta}^* \mathcal{X}{'})\}. 
 \vspace{-0.1in}
\end{align}

\noindent \textbf{Time complexity} of our appliance recognition model mainly depends on the operations at the convolutional layers. As the model comprises three layers with identical configurations, the time complexity is $O\big (m(\sum_{i=1}^{3} p_{i-1} \cdot s^2_{i} \cdot f_{i} \cdot o_{i}^2)\big)$~\cite{he2015convolutional}, where $m$ is the length of the instance (i.e., appliance footprint), $p_{i-1}$ is the number of input channels, $s_{i}$ and $f_{i}$ are  respectively the number and size of filters, and $o_i$ is the spatial size of output feature vector at $i^{th}$ layer. As the model is trained across different clients, which can be done in parallel, FedAR+ imposes only communication overheads due to aggregation after each FL round, however, the communication analysis is out-of-scope of this work.   

\vspace{-0.1in}
\section{Performance Evaluation}\label{results}
We evaluate the performance of FedAR+ on a dataset collected from three different houses, and also on two existing datasets, namely UK-DALE~\cite{UK-DALE} and Tracebase~\cite{6388037}, using accuracy, precision, recall, and $F_1$-score. To demonstrate the superiority of our proposed approach, we compare it with three existing ones and report the results using the considered metrics and execution time.

\vspace{-0.2cm}
\subsection{Datasets}
\subsubsection{Collected dataset} Data collection is done by deploying appliance-specific smart plugs in three different houses (not included in the auxiliary dataset) for a period of one month. The power consumption data are collected at a sampling rate $1$Hz. After preprocessing (discussed in Section~\ref{data_prep}), we get a dataset of $840$ instances for six different appliances including refrigerator, microwave oven, television, washing machine, air conditioner, and mixer grinder. Each instance corresponds to a footprint of a particular appliance. We call this dataset as {\em appliance footprints in residential buildings} (Res-AF). 

\subsubsection{UK-DALE dataset~\cite{UK-DALE}} It contains power consumption data of various household appliances from five houses for a period of one year. The data are recorded at a sampling rate $1/6$ Hz (i.e., one reading per six seconds). For the experiments, we selected five appliances including refrigerator, washing machine, kettle, dishwasher, and boiler.

\subsubsection{Tracebase dataset~\cite{6388037}} It contains more than $1000$ power consumption traces collected from $15$ different houses. One trace corresponds to a time series of readings taken for one particular appliance over a window of $24$ hours. The readings are reported at an average sampling rate of $1$ Hz. In~\cite{6388037}, a Measurement and Actuation Unit (MAU) is developed to collect the power consumption traces of the appliances. MAU is installed between wall mounted power outlet and power plug of the appliance. For experimental evaluation, we selected five appliances having sufficient number of traces (instances) in the dataset. The five selected appliances are refrigerator, microwave oven, kettle,  television, and dishwasher.    

\vspace{3pt}
\noindent $\bullet$ \textbf{Preprocessing:} In the existing datasets, the readings are collected for a continuous period and thus the resulting time series include both ON and OFF states of the appliances. We therefore preprocess these datasets to prepare them for training and testing the proposed recognition model. By following the preprocessing steps described in Section~\ref{data_prep}, we obtained $1,860$ and $930$ instances (i.e., appliance footprints) in UK-DALE and Tracebase datasets, respectively.


 \vspace{-0.1in}
 \subsection{Experimental setup}
  Prior to conducting experiments, we split each dataset into two parts: training with $80\%$ and testing with $20\%$ instances. The training data are further split into $10$ non-iid chunks using Dirichlet distribution with parameter $\alpha=0.9$ and $20\%$ overlapping. These chunks are then provided to $K=10$ clients.  It is to note that each client may have different number of instances per class due to non-iid data. While working with Res-AF dataset, a copy of auxiliary dataset (collected in Section~\ref{prelim}) is given to each client. Similarly, for UK-DALE and Tracebase datasets, $20\%$ of the training data are used as auxiliary dataset which is same for all clients and required for the model initialization. Further, to produce noisy labels in the training datasets, we flip the labels of a fixed percentage (indicated in the respective results) of instances across all class labels.

We simulated the FedAR+ algorithm with a server and $10$ clients in Python programming language through Tensorflow libraries. In the implementation, we chose the following parameters: optimizer = `sgd', activation = `relu' with each convolutional layer, and learning rate $\eta=0.1$. The number of FL global rounds are set based on the results obtained after rigorous experiments (see Section~\ref{result1}). Since our recognition model consists of only three convolutional layers, each with $128$ filters, it does not overfit with small training datasets.     

\vspace{-0.1in}
\subsection{Performance metrics}
The following metrics evaluate the performance of FedAR+.

\begin{itemize}
  \item \textit{Precision} (P): It is the ratio of the number of correctly classified instances of an appliance $x$ to the total number of instances classified as $x$. Precision indicates a quality aspect of the appliance recognition model.      
 \item \textit{Recall} (R): It is the ratio of the number of correctly classified instances of an appliance $x$ to the total number of instances actually belonging to $x$. Recall measures the completeness and relevance of the recognition model.
 \item \textit{F$_1$ score:} It is a harmonic mean of precision and recall, and is computed as $\frac{2 \times \text{P} \times \text{R}}{\text{P}+\text{R}}$.

\item \textit{Accuracy:} It is the proportion of correctly classified instances of the testing dataset. It is expressed as

\vspace{-0.1in}
\begin{align}\nonumber
 \text{Accuracy} =\frac{1}{N_{test}} \sum_{i=1}^{N_{test}} \mathbb{I}(\mathcal{Y}^i, \mathcal{Y}{'}), 
\end{align}
\vspace{-0.1in}

\noindent
where $\mathbb{I}(\mathcal{Y}^i, \mathcal{Y}{'}) = 1$ if $\mathcal{Y}^i==\mathcal{Y}{'}$ and $0$ otherwise; and $N_{test}$ is the number of instances in testing dataset. 
\vspace{-0.05in}
\end{itemize}

\subsection{Experimental results}
Through experiments, we seek answers to the following six questions:
 (1) How does the accuracy of the model improve over rounds in FedAR+?
 (2) How does the concentration of noisy labels influence the testing performance? 
 (3) What is the appliance-wise performance of the model? 
 (4) How does FedAR+ scale to the number of clients?
 (5) How efficiently does FedAR+ outperform the prior approaches? 
The results are presented below.

\subsubsection{Training accuracy over FL rounds} \label{result1}
\begin{figure}[h]
\centering
\minipage{0.235\textwidth}
\centering
    \includegraphics[scale=0.42]{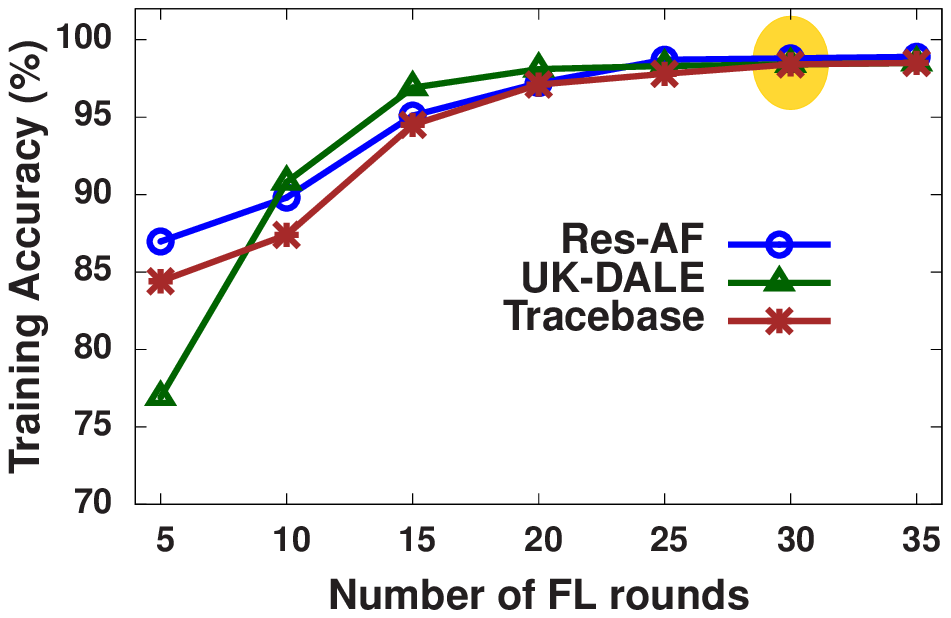}
    \subcaption{\footnotesize{With $5\%$ noisy labels}}
\endminipage\hfill
\minipage{0.235\textwidth}
\centering
    \includegraphics[scale=0.42]{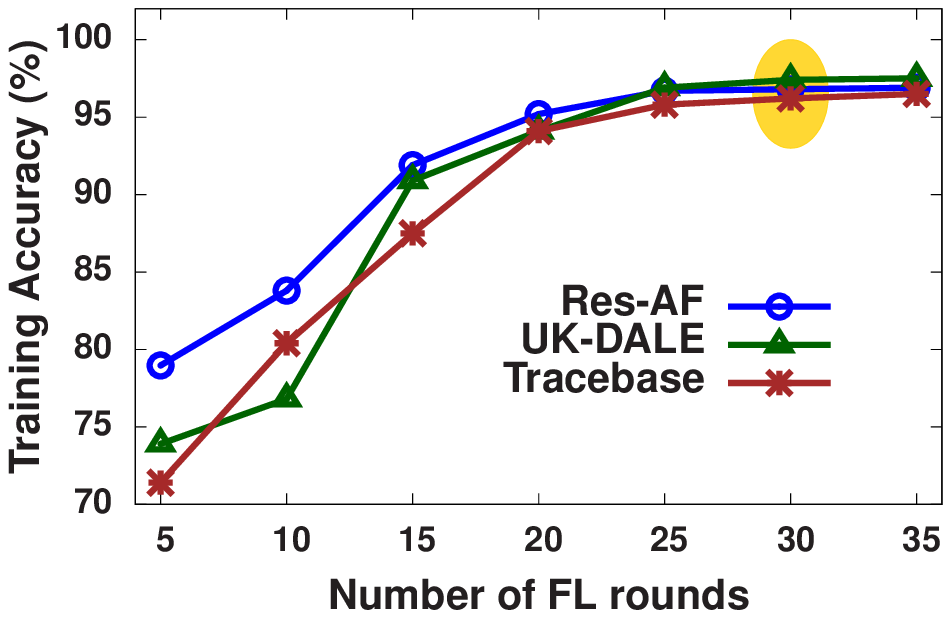}
    \subcaption{\footnotesize{With $30\%$ noisy labels}}
\endminipage\hfill
\vspace{-0.1in}
 \caption{Training accuracy of the proposed appliance recognition model with the noise handling method in FedAR+.}
\label{epoch_setting}
\vspace{-0.1in}
\end{figure}

At first, we analyze the training accuracy of the recognition model over $35$ rounds with different concentrations (from $5\%$ to $30\%$) of noisy labels in the training dataset. We set the local epochs to $50$ at the clients in all the experiments. Figure~\ref{epoch_setting} demonstrates the results for $5\%$ and $30\%$ cases. 
The results clearly indicate that the model is able to achieve more than $92\%$ of training accuracy at $30^{th}$ round, even when $30\%$ training instances are mislabeled. 
As initial model is far from the optimality in first few rounds, it shows low accuracy for all datasets. The accuracy increases rapidly up to $15^{th}$ rounds and starts stabilizing afterwards. As no change is observed in the accuracy between $30^{th}$ and $35^{th}$ rounds, we report all the subsequent results with $30$ FL rounds and $50$ local epochs at the clients.

 

\subsubsection{Testing performance with varying concentration of noisy labels} \label{result2}
\begin{figure}[h]
 \vspace{-0.1in}
\centering
\minipage{0.235\textwidth}
\centering
   \includegraphics[scale=0.42]{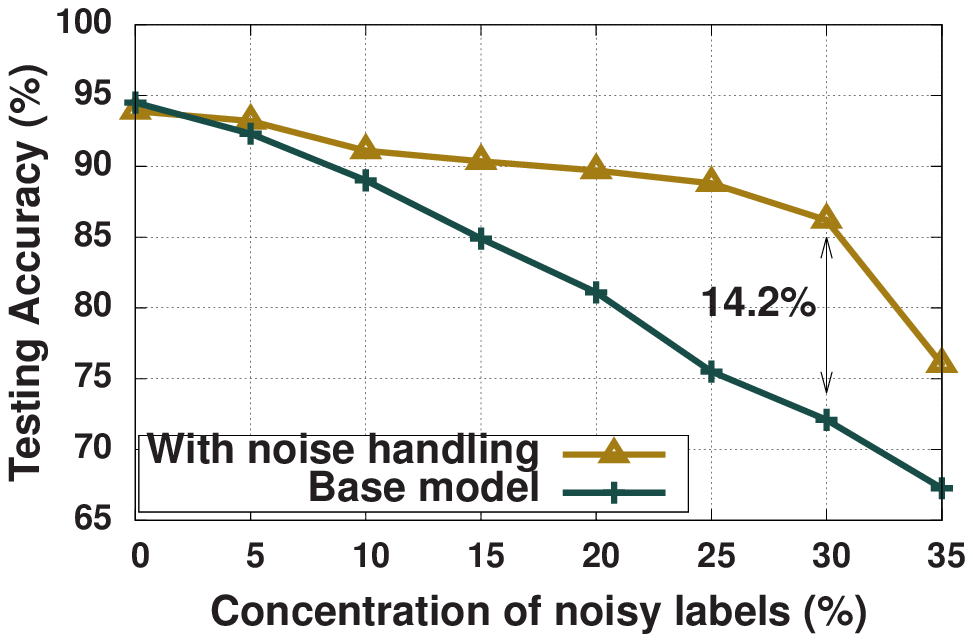}
    \subcaption{\footnotesize{Accuracy for Res-AF dataset}}
\endminipage\hfill
\centering
\minipage{0.235\textwidth}
\centering
    \includegraphics[scale=0.42]{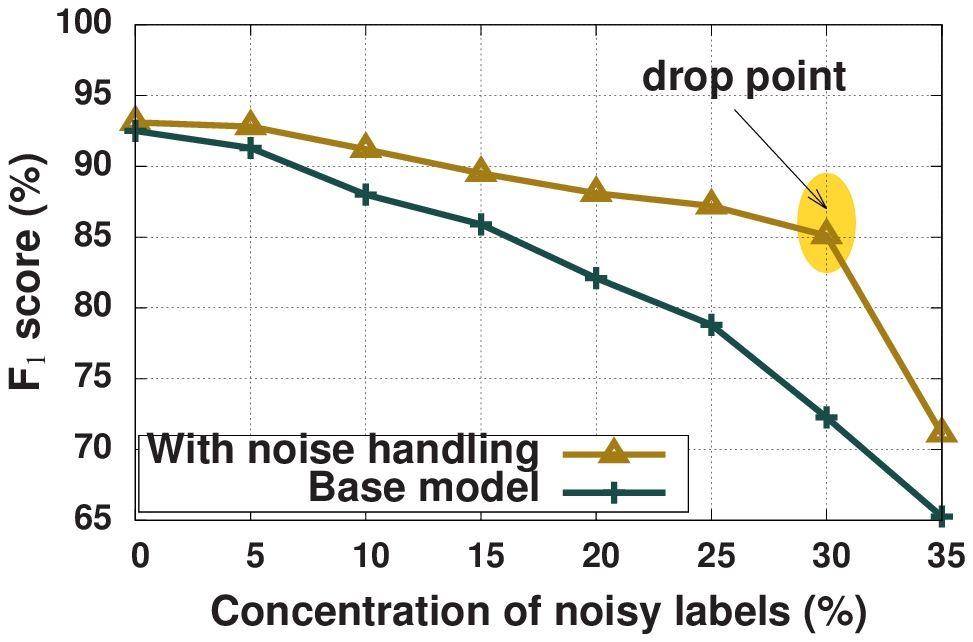}
    \subcaption{\footnotesize{F$_1$ score for Res-AF dataset}}
\endminipage\hfill
\minipage{0.235\textwidth}
\centering
    \includegraphics[scale=0.42]{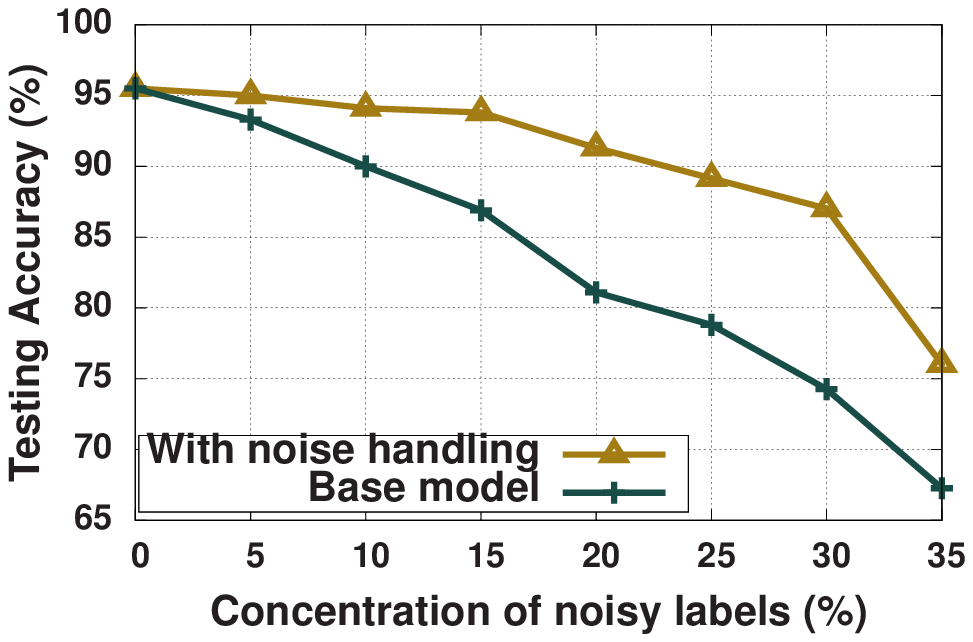}
    \subcaption{\footnotesize{Accuracy for UK-DALE dataset}}
\endminipage\hfill
\centering
\minipage{0.235\textwidth}
\centering
    \includegraphics[scale=0.42]{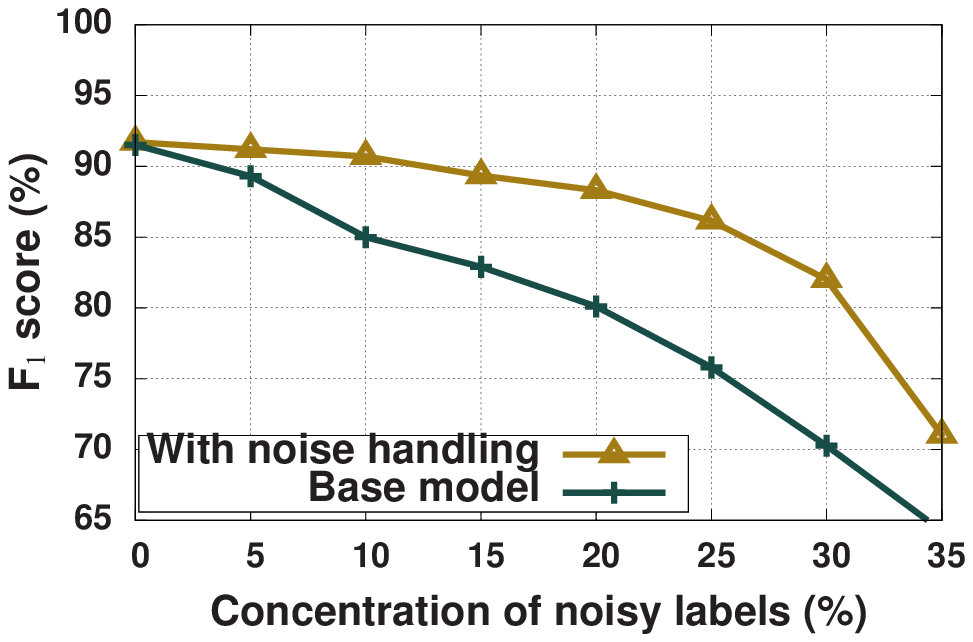}
    \subcaption{\footnotesize{F$_1$ score for UK-DALE dataset}}
\endminipage\hfill
\minipage{0.235\textwidth}
\centering
   \includegraphics[scale=0.42]{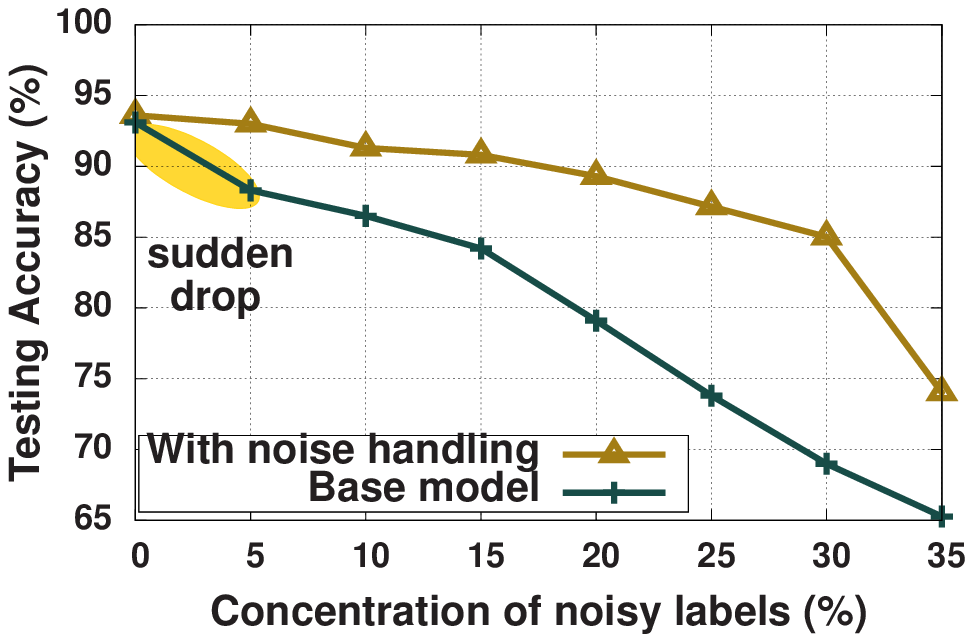}
    \subcaption{\footnotesize{Accuracy for Tracebase dataset}}
\endminipage\hfill
\minipage{0.235\textwidth}
\centering
    \includegraphics[scale=0.42]{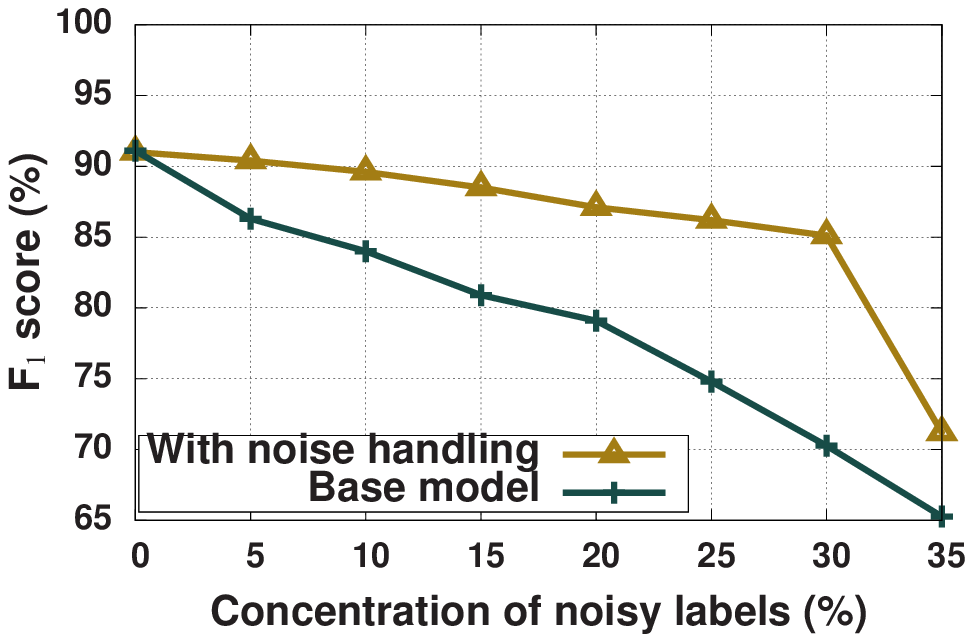}
    \subcaption{\footnotesize{F$_1$ score for Tracebase dataset}}
\endminipage\hfill
\vspace{-0.1in}
\caption{Performance results of FedAR+ using the recognition model with and without noise handling method.} 
\label{noisy}
\vspace{-0.1in}
\end{figure}

In Figure~\ref{noisy}, we report the impact of noisy labels with varying concentrations on the performance of the model. As the concentration of noisy labels increases, the accuracy and F$_1$ score decrease, which seems bit obvious but, such a drop is substantial for the base model compared to the one with noise handling method. For instance, in case of Tracebase dataset, with just $5\%$ noisy labels, the base model immediately loses more than $4\%$ accuracy. On the flip side, for Res-AF dataset with $30\%$ noisy labels, the model gains $14.2\%$ on accuracy by utilizing the proposed noise handling method, indicating the effectiveness of the proposed approach. For all the datasets, FedAR+ achieves an accuracy of more than $84\%$ and F$_1$ score of above $81\%$ up to $30\%$ of noisy labels; however, the performance drops sharply afterwards, signaling the noise handling upper limit of our approach.
With higher concentrations, the reason for such a drop is the increase in confusion while differentiating between correct and noisy labels. Similar observations can be made from F$_1$ score, shown in parts (b), (d) and (f) of Figure~\ref{noisy}.    
 
\subsubsection{Appliance-wise performance of the recognition model} \label{result3}
Next, we analyze the appliance-wise performance results of the model with noise handling method for Res-AF and Tracebase datasets in Tables~\ref{tfa_p} and~\ref{tracebase_p}, respectively. The results are reported using precision, recall, and F$_1$ score evaluation metrics for $5\%$ and $30\%$ concentration of noisy labels. Results indicate that the model performs best on ``television'' and ``refrigerator'' classes; they were identified with more than $90\%$ of recall even when the concentration of noisy labels is $30\%$, witnessing the robustness of our approach FedAR+ against noisy labels. 
We also observed that the precision values are marginally ($1\sim4$ approximately) differ from the recall ones, indicating the ability of FedAR+ to manage the good balance between the relevance and completeness of the appliance recognition model. 

\begin{table}[h]
\centering
\small
\caption{Appliance-wise performance of the model with noisy handling method for \textit{Res-AF} dataset using precision (P), recall (R), and F$_1$ score (F).}
\vspace{-0.1in}
\resizebox{0.48\textwidth}{!}{
\begin{tabular}{|c|c|c|c|c|c|c|}
\hline
\multirow{2}{*}{}        & \multicolumn{3}{c|}{\textbf{5\% Noisy labels}}               & \multicolumn{3}{c|}{\textbf{30\% Noisy labels}}              \\ \cline{2-7} 
                         & P (\%) & R (\%) & F (\%) & P (\%) & R (\%) & F (\%) \\ \hline
Refrigerator    &       90.4          &      94.0           &        90.8         &      90.2          &         89.8        &      90.0           \\ \hline
Microwave oven  &         92.8        &       92.7          &     93.5            &       85.3     &         85.5        &     85.4           \\ \hline
Television  &        93.1        &       94.4      &        94.7         &          90.5       &        90.8       &       90.6            \\ \hline
Washing machine  &     89.0            &     91.4            &        89.1         &      87.2           &       80.4          &      87.8           \\ \hline
Air conditioner  &       89.3          &          92.7       &          90.7       &         80.1        &        82.4         &     81.2            \\ \hline
Mixer grinder  &        91.3         &        90.0         &          90.5       &          83.8       &         85.2        &              84.5   \\ \hline
\textbf{Average}  &         90.8        &      92.5           &     91.5            &       85.8          &          87.3       &    86.5             \\ \hline
\end{tabular}
}
\label{tfa_p}
\end{table}

\begin{table}[ht!]
\vspace{-0.1in}
\centering
\small
\caption{Appliance-wise performance of the model with noise handling method for \textit{Tracebase} dataset using precision (P), recall (R), and F$_1$ score (F).}
\vspace{-0.1in}
\resizebox{0.48\textwidth}{!}{
\begin{tabular}{|c|c|c|c|c|c|c|}
\hline
\multirow{2}{*}{}        & \multicolumn{3}{c|}{\textbf{5\% Noisy labels}}               & \multicolumn{3}{c|}{\textbf{30\% Noisy labels}}              \\ \cline{2-7} 
                         & P (\%) & R (\%) & F (\%) & P (\%) & R (\%) & F (\%) \\ \hline
Refrigerator    &       91.1         &        94.2         &        92.6    &          86.3       &       91.5        &    88.6     \\ \hline
Microwave oven  &      88.4       &      91.2           &        89.7        &          81.2        &        89.1        &     84.9    \\ \hline  
Kettle  &          86.7        &       88.2          &       87.4         &           80.3      &        85.2         &       82.6          \\ \hline
Television  &         88.1         &        93.2         &          90.5 &         85.3        &        92.1         &    88.5    \\ \hline 
Dishwasher  &         87.2        &         90.1         &        88.6        &      81.1           &     87.9            &   84.3              \\ \hline
\textbf{Average}  &       88.3          &         91.4        &       89.8          &         82.8        &       89.2        &        85.7          \\ \hline
\end{tabular}
}
\label{tracebase_p}
\vspace{-0.1in}
\end{table}

\subsubsection{Scalability analysis}
\begin{figure}[t]
\centering
\minipage{0.235\textwidth}
\centering
    \includegraphics[scale=0.42]{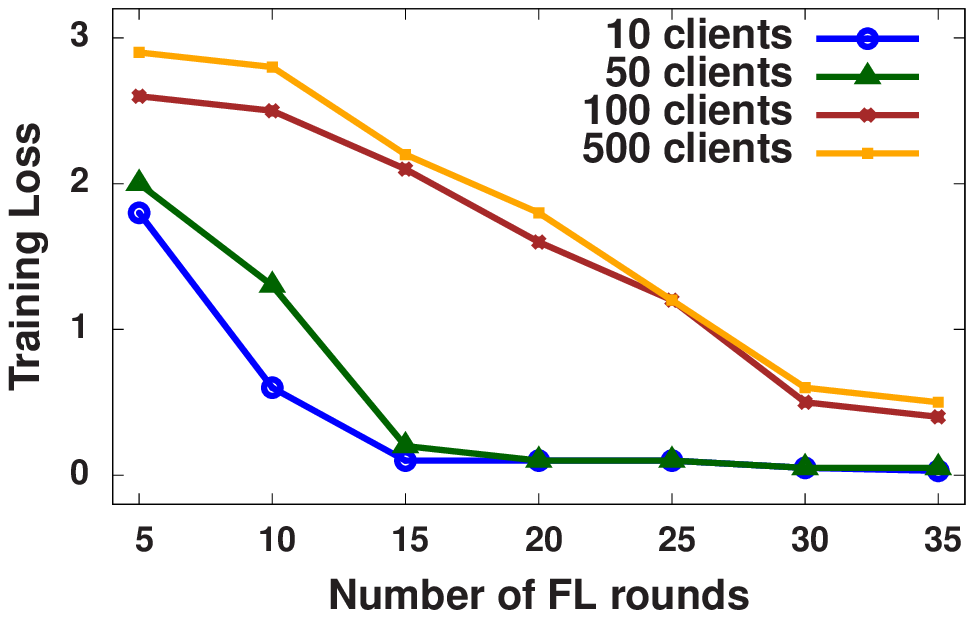}
    \subcaption{\footnotesize{For Res-AF dataset}}
\endminipage\hfill
\minipage{0.235\textwidth}
\centering
    \includegraphics[scale=0.42]{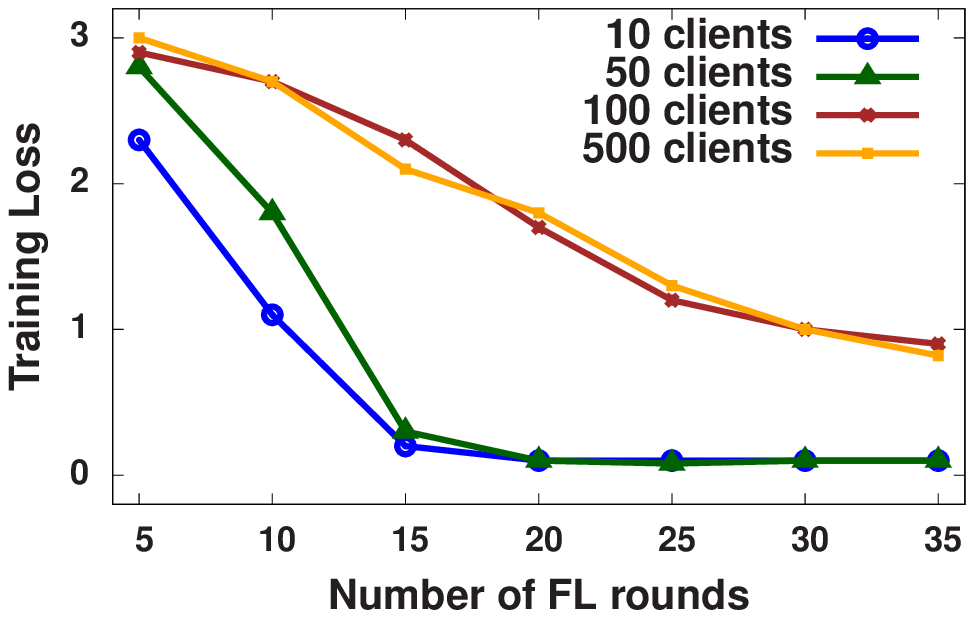}
    \subcaption{\footnotesize{For Tracebase dataset}}
\endminipage\hfill
\vspace{-0.1in}
 \caption{Training loss of the model in FedAR+ with $0\%$ noisy labels.}
\label{scale}
\vspace{-0.1in}
\end{figure}

The scalability of our FedAR+ algorithm can be measured in terms of the number of clients it can support without affecting the performance of the model. We investigate the scalability by increasing the number of clients. To perform an experiment with many clients (say $50\sim500$), we first need to provide sufficient data instances to every client. For this, we generated synthetic training and testing data by adding Gaussian noise with $\mu=0$ and $\sigma \in [0.1 - 0.5]$, and distributed the data uniformly among the clients. Figure~\ref{scale} shows the training loss of the global model over FL rounds for different number of clients. One quick observation from the results is that the model converges (i.e., loss stabilizes) after $20$ rounds when only $50$ clients exist, however it needs $10$ more rounds with $100$ to $500$ clients. It indicates that by increasing the number of FL rounds, our approach can be easily scaled to large number of clients without affecting the convergence. Furthermore, we conducted some experiments with $0\%$ and $30\%$ noisy labels and reported the accuracy (obtained by the model trained for $30$ rounds) in Table~\ref{scale1} for both Res-AF and Tracebase datasets. The results demonstrate the scalability of FedAR+ to $500$ clients with a marginal drop in accuracy that cab be easily recovered by training the model for more FL rounds.

\begin{table}[h]
\vspace{-0.1in}
\centering
\caption{Accuracy results for the model trained over $30$ rounds.}
\vspace{-0.1in}
\resizebox{0.4\textwidth}{!}{
\begin{tabular}{|c|cc|cc|}
\hline
\multirow{2}{*}{\textbf{\begin{tabular}[c]{@{}c@{}}No. of\\ Clients\end{tabular}}} & \multicolumn{2}{c|}{\textbf{Res-AF dataset}}                                                                                                     & \multicolumn{2}{c|}{\textbf{Tracebase dataset}}                                                                                               \\ \cline{2-5} 
                                                                                   & \multicolumn{1}{c|}{\begin{tabular}[c]{@{}c@{}}0\% Noisy\\ labels\end{tabular}} & \begin{tabular}[c]{@{}c@{}}30\% Noisy \\ labels\end{tabular} & \multicolumn{1}{c|}{\begin{tabular}[c]{@{}c@{}}0\% Noisy\\ labels\end{tabular}} & \begin{tabular}[c]{@{}c@{}}30\% Noisy\\ labels\end{tabular} \\ \hline
10                                                                                 & \multicolumn{1}{c|}{93.9}                                                       & 88.1                                                         & \multicolumn{1}{c|}{93.5}                                                       & 87.3                                                     \\ \hline
50                                                                                 & \multicolumn{1}{c|}{94.1}                                                       & 87.9                                                         & \multicolumn{1}{c|}{92.1}                                                       & 86.2                                                        \\ \hline
100                                                                                & \multicolumn{1}{c|}{91.8}                                                       & 85.2                                                         & \multicolumn{1}{c|}{91.2}                                                       & 84.8                                                        \\ \hline
500                                                                                & \multicolumn{1}{c|}{92.5}                                                       & 84.7                                                         & \multicolumn{1}{c|}{91.1}                                                       & 83.3                                                        \\ \hline
\end{tabular}
}
\label{scale1}
\vspace{-0.1in}
\end{table}

\setlength\doublerulesep{0.2pt}
\begin{table*}[h]
\centering
\footnotesize
\caption{Performance comparison of the proposed approach with the existing ones using accuracy (in \%). [\textbf{P}: Precision, \textbf{R}: Recall, \textbf{F}: $F_1$ score, \textbf{A}: Accuracy]}
\vspace{-0.1in}
\resizebox{1\textwidth}{!}{
\begin{tabular}{|c|c|c|c|c|c||c|c|c|c||c|c|c|c||c|c|c|c|}
\hline
\multirow{3}{*}{\textbf{Datasets}} & \multirow{3}{*}{\textbf{\begin{tabular}[c]{@{}c@{}}Approaches\end{tabular}}} & \multicolumn{16}{c|}{\textbf{Concentration of noisy labels in training data}}                         \\ \cline{3-18} 
                                            & & \multicolumn{4}{c|}{$0\%$} & \multicolumn{4}{c|}{$10\%$}   & \multicolumn{4}{c|}{$20\%$}   & \multicolumn{4}{c|}{$30\%$} \\ \cline{3-18} 
                                            & & \textbf{P} & \textbf{R} & \textbf{F} & \textbf{A} & \textbf{P} & \textbf{R} & \textbf{F} &\textbf{A} & \textbf{P} & \textbf{R} & \textbf{F} & \textbf{A} & \textbf{P} & \textbf{R} & \textbf{F} & \textbf{A} \\ \hline 
\multirow{5}{*}{Res-AF}  &  HARB~\cite{yan2019household}                & $84.2$ & $87.2$  & $85.7$ &   $89.5$   & $81.2$ & $83.1$ & $82.1$ &    $84.9$   & $72.4$ & $67.2$ & $69.7$ &     $70.2$   & $56.1$ &  $60.1$ & $58.0$ &   $58.6$ \\ \cline{2-18}
                     &  LSTM-AR~\cite{schwermer2022federated}               & $60.2$ & $59.2$ & $59.7$  & $60.5$   & $59.2$ & $55.2$ & $57.1$ &  $60.9$   & $52.2$ & $50.2$ & $50.7$ &     $51.1$     & $43.2$ & $45.7$ & $44.4$ &  $46.6$ \\ \cline{2-18}
                     & CNN-AR~\cite{schwermer2022federated}                &  $55.2$  & $53.4$ & $54.2$ & $57.3$   & $52.4$ & $52.3$ & $52.3$ &  $50.8$   & $45.1$ & $44.1$ & $43.6$ &  $47.3$    & $48.2$ &  $41.1$ & $42.1$ &   $41.7$   \\ \cline{2-18}   
                     & FedAR (Proposed)                                       & $92.4$ & $93.8$  & $93.1$ &  $94.4$     & $90.1$ & $87.5$ & $88.8$ &  $89.1$    &   $82.1$  & $79.5$ & $80.8$ & $81.5$     & $73.6$ & $72.4$ & $73.0$  &  $73.4$     \\ \cline{2-18}
                     & FedAR+ (Proposed) & $\mathbf{93.1}$ & $\mathbf{95.1}$ & $\mathbf{94.1}$ &  $\mathbf{94.8}$      & $\mathbf{91.2}$ & $\mathbf{89.5}$ & $\mathbf{90.3}$ &  $\mathbf{92.3}$   & $\mathbf{88.5}$ & $\mathbf{87.1}$ & $\mathbf{87.8}$ &   $\mathbf{89.5}$      & $\mathbf{87.2}$ & $\mathbf{85.7}$ & $\mathbf{86.4}$ &  $\mathbf{88.4}$      \\ \hhline{==================}
\multirow{5}{*}{UK-DALE} &  HARB~\cite{yan2019household}              & $87.6$ &  $89.5$ & $88.5$ &  $90.5$   & $85.7$ & $81.2$ & $83.4$ &  $83.5$     & $70.2$ & $68.6$ & $69.4$ &     $72.5$     & $60.2$ & $56.1$ & $58.1$ &  $59.2$  \\ \cline{2-18}
                     &  LSTM-AR~\cite{schwermer2022federated}              & $59.1$ & $58.9$  & $59.0$ &  $60.5$   & $56.3$ & $59.4$ & $57.8$ &  $58.9$   & $45.6$ & $47.3$  & $45.9$ &    $43.2$      & $43.2$ & $41.1$ & $42.1$ &  $43.6$ \\ \cline{2-18}
                     & CNN-AR~\cite{schwermer2022federated}              & $56.1$ & $50.1$ & $53.0$   & $54.8$   & $50.6$ & $59.2$ & $59.9$ &  $51.2$   & $43.5$ & $43.2$ & $41.3$  &   $42.2$   & $36.4$ & $38.1$ & $37.7$ & $41.9$     \\ \cline{2-18}   
                     & FedAR (Proposed)                                        & $92.4$ & $91.2$  &  $91.8$  &   $94.9$   & $87.2$ & $89.2$ & $88.2$  & $90.3$    & $75.2$ & $76.1$ & $75.6$ &  $81.1$      & $76.2$ & $70.3$ &  $73.1$ & $74.2$    \\ \cline{2-18}
                     & FedAR+ (Proposed)   & $\mathbf{94.1}$ &  $\mathbf{90.5}$ & $\mathbf{92.3}$  &    $\mathbf{95.1}$   & $\mathbf{94.8}$ & $\mathbf{90.2}$ & $\mathbf{92.4}$ &  $\mathbf{93.5}$   & $\mathbf{86.4}$ & $\mathbf{87.7}$ & $\mathbf{87.0}$ &  $\mathbf{90.8}$      & $\mathbf{84.2}$ & $\mathbf{80.9}$ & $\mathbf{82.5}$ &  $\mathbf{87.2}$      \\ \hhline{==================}
\multirow{5}{*}{Tracebase} &  HARB~\cite{yan2019household}               & $87.4$ & $88.1$ & $87.7$ &  $90.8$    &  $78.6$ & $82.1$ & $80.3$ &  $80.3$     & $71.5$ & $68.2$ & $69.8$  &  $70.3$      & $58.2$ & $52.1$ & $55.0$ & $57.4$  \\ \cline{2-18}
                     &  LSTM-AR~\cite{schwermer2022federated}                & $56.2$ & $60.3$ & $58.2$  & $59.5$    & $58.2$ & $60.2$ & $59.2$ &  $59.9$     & $50.5$ & $51.2$ & $49.3$ &   $51.2$     & $45.2$ & $45.4$ & $45.0$ &  $42.6$ \\ \cline{2-18}
                     & CNN-AR~\cite{schwermer2022federated}                & $49.2$ & $50.8$ & $50.0$  &  $51.2$   & $48.7$ & $52.4$ & $50.5$ &   $50.5$   & $43.6$ & $38.7$ & $41.1$ &   $43.3$     & $37.4$ & $36.9$ & $40.1$ & $39.2$    \\ \cline{2-18}   
                     & FedAR (Proposed)                                      & $94.2$ & $92.8$  & $\mathbf{93.5}$ &     $93.5$    & $91.7$ & $90.7$ & $91.2$ &    $90.2$   & $80.5$ & $78.7$ & $79.6$ &      $80.8$    & $70.1$& $71.9$ & $71.0$ &   $72.7$      \\ \cline{2-18}
                     & FedAR+ (Proposed)   & $\mathbf{94.6}$ & $\mathbf{91.4}$ & $\mathbf{93.0}$ &  $\mathbf{93.8}$   & $\mathbf{93.7}$ & $\mathbf{90.5}$ & $\mathbf{92.1}$ &    $\mathbf{92.9}$    & $\mathbf{89.2}$ & $\mathbf{86.3}$ & $\mathbf{87.7}$ &   $\mathbf{88.9}$       & $\mathbf{87.5}$ & $\mathbf{84.3}$ & $\mathbf{85.9}$  & $\mathbf{86.1}$   \\ \hline                     
\end{tabular}
}
\label{comp1}
\vspace{-0.1in}
\end{table*}

\vspace{-0.1in}
\subsection{Comparison with existing approaches} \label{result5}
We compare FedAR+ approach with three state-of-the-art solutions including two best performing plug-load identification models from~\cite{schwermer2022federated} and Household Appliance Recognition through Bayes classification (HARB)~\cite{yan2019household}. The work~\cite{schwermer2022federated} leverages FL to train four different deep learning models; we pick two best performers, long-short term memory (LSTM) and convolutional neural network (CNN), named as LSTM-AR and CNN-AR for the convenience. Similar to the proposed approach, LSTM-AR and CNN-AR models are also trained across $10$ clients (possessing non-iid data) over $30$ FL rounds with aggregation at every $50$ local epochs. On the other hand, HARB~\cite{yan2019household} follows a central learning paradigm. In HARB, the time series data are transformed into a set of statistical features (e.g., working time, power range, frequency of the use, etc.) on which Bayesian learning is applied to obtain posterior class probabilities that are used for appliance prediction. As the existing solutions do not incorporate any noise handling method, we consider two variants of the proposed approach: 1) FedAR: without noise handling and 2) FedAR+: with noise handling, in the interest of fair comparison. Table~\ref{comp1} shows the comparison results using precision, recall, $F_1$ score, and  accuracy. We make following important observations:
\begin{itemize}
\item Both FedAR and FedAR+ gain over the existing solutions on all the evaluation metrics by a large margin (approximately $15\%\sim27\% $) even in the presence of $30\%$ noisy labels.
\item Prior FL models, LSTM-AR and CNN-AR, perform much worse than the centralized algorithm HARB in all the cases, indicating their inability to learn with non-iid data across the clients. On the contrary, our FL approach FedAR outperforms HARB with a substantial margin of more than $5\%$ in case of no noisy label and more than $12\%$ in case of $30\%$ noisy labels, showing the effectiveness of learning with non-iid data with the aggregation function (defined in Eq.~\ref{agg}). 

 \item Even with $30\%$ mislabeled data, FedAR+ secured the accuracy and F$_1$ of more than $85\%$ on all the datasets, validating the success of our noise handling method. It is worth to notice that the performance gain of FedAR+ over other methods increases significantly with the surge in noisy labels.

\end{itemize}

\subsubsection{IID versus non-IID data}
\begin{figure}[h]
\centering
\minipage{0.235\textwidth}
\centering
   \includegraphics[scale=0.4]{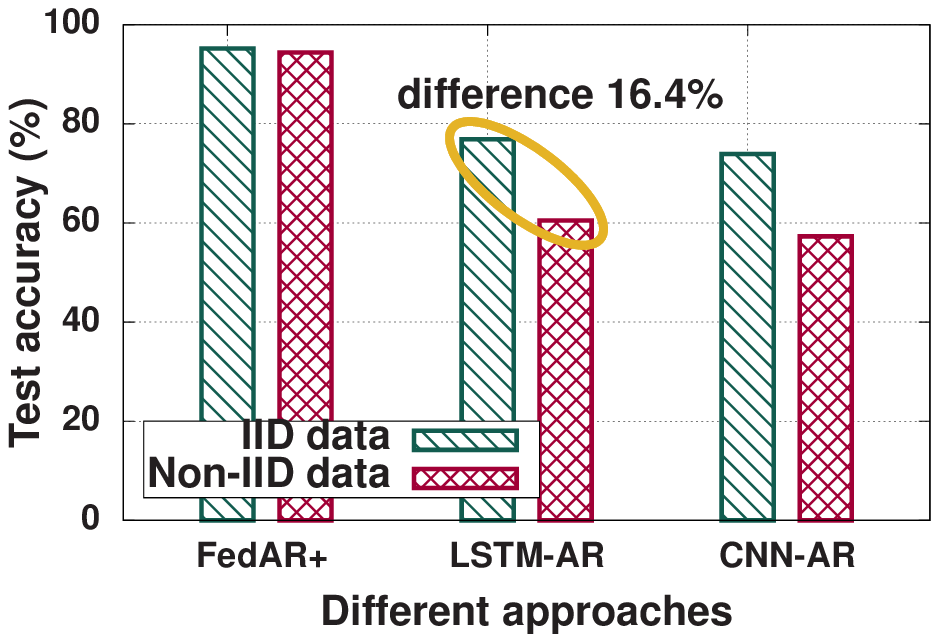}
    \subcaption{\footnotesize{For Res-AF dataset}}
\endminipage\hfill
\minipage{0.235\textwidth}
\centering
    \includegraphics[scale=0.4]{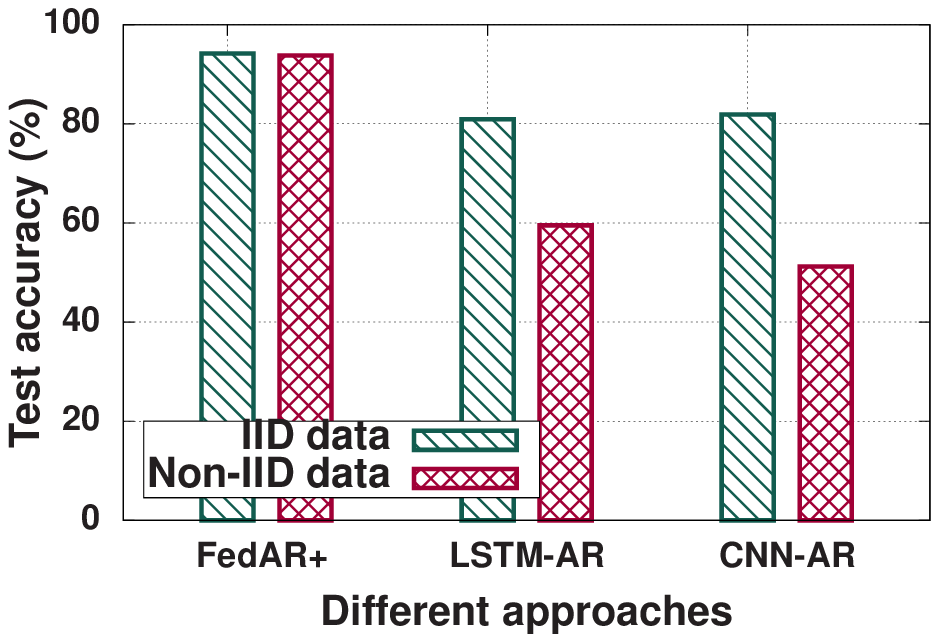}
    \subcaption{\footnotesize{For Tracebase dataset}}
\endminipage\hfill

\vspace{-0.1in}
\caption{Comparing accuracy results obtained with iid and non-iid data (with $0\%$ noisy labels) across clients in FL based approaches.  }
\label{data_dist}
\end{figure}

Considering $10$ clients in FL setup, we now report the test accuracy results in Figure~\ref{data_dist} with iid (or uniformly) and non-iid data (simulated using Dirichlet distribution as mentioned in experimental setup). For the sake of fair comparison, this experiment does not include noisy labels. For both datasets, FedAR+ shows its capability to learn with non-iid data by securing almost equal accuracy as with iid data, however it is not true for prior approaches; for instance, LSTM-AR loses $16.4\%$ accuracy when clients possess non-iid training data. Although the performance of prior approaches seem to improve significantly with iid data, they could never reach beyond $81\%$.            
 
\subsubsection{Execution time} 
\begin{figure}[h]
\centering
\minipage{0.235\textwidth}
\centering
   \includegraphics[scale=0.4]{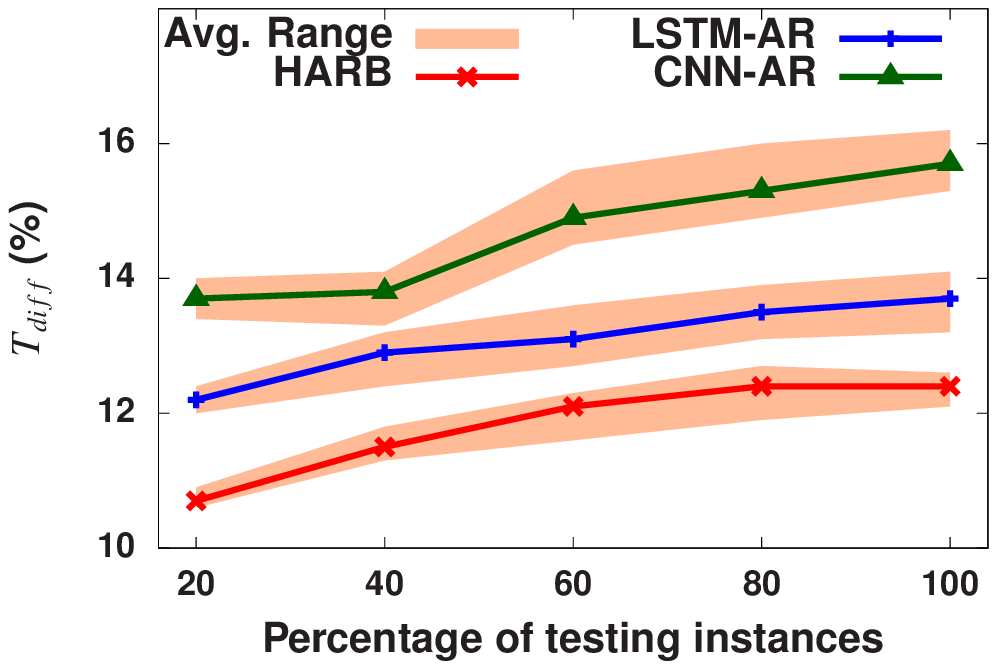}
    \subcaption{\footnotesize{For Res-AF dataset}}
\endminipage\hfill
\minipage{0.235\textwidth}
\centering
    \includegraphics[scale=0.4]{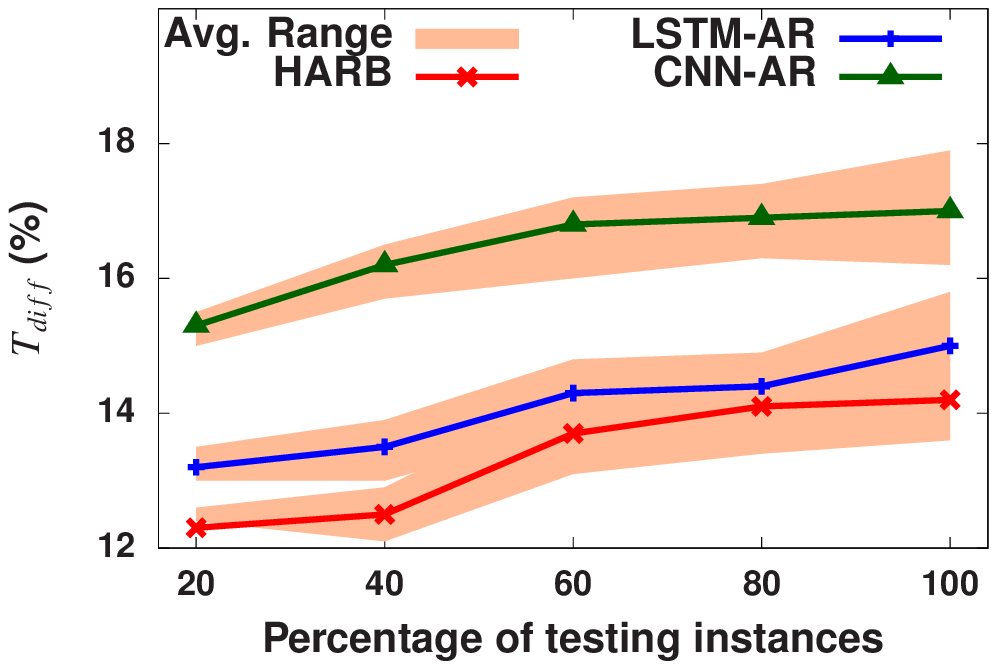}
    \subcaption{\footnotesize{For Tracebase dataset}}
\endminipage\hfill
\vspace{-0.1in}
\caption{Comparison results using percentage difference in the execution time of the existing approaches from FedAR+.}
\label{execution}
\end{figure}

Finally, we compare the execution time of FedAR+ with the existing approaches. Here, the execution time indicates the total time taken by an approach to classify the entire testing dataset. To better understand the comparison, we compute the percentage difference ($T_{diff}$) in the execution times of any existing approach from FedAR+, as follows:   
\begin{align}\nonumber
T_{diff} =\frac{T_{x} - T_{FedAR+}}{T_x} \times 100,
\end{align}
where $T_{x}$ denotes the execution time of an existing approach $x$. Figure~\ref{execution} shows the comparison results in terms of $T_{diff}$, which is an average over 50 executions. It is apparent that $T_{diff}$ is positive in all the cases, indicating that FedAR+ is faster ($10.5\%$ to $16.8\%$) than the existing approaches. 
 
\vspace{-0.1in}
\section{Conclusion and Discussions}\label{conclude}
This paper proposed an FL approach, FedAR+, for identifying household appliances using their electricity consumption patterns. The approach dealt with two important issues related to the appliance recognition model: 1) presence of noisy labels in the training dataset, and 2) model building at the client (consumer) side without sharing local data to the server. By employing deep learning and incorporating a noise handling method, we developed an accurate appliance recognition model that can learn from mislabeled training data. To preserve client's privacy, the model is trained and updated locally without sharing the local data with the server. By deploying smart plugs in an apartment complex, we collected a real world dataset to validate the effectiveness of FedAR+. Through rigorous experimental analysis, we demonstrated the superiority of our approach over existing ones and showed that FedAR+ can effectively accommodate up to $30\%$ noisy labels while compromising the accuracy only slightly. Considering the availability of sufficient training data, FedAR+ can be scaled to large number of clients, enabling its adoption to real-world energy monitoring applications. 

In future, we plan to work on theoretical guarantees of FedAR+. We will also explore the robustness aspects of the appliance recognition model under the presence of alien and malicious clients. 



{\em Alien appliance:} An alien (unseen) appliance is one for which there exists no instance in the training dataset to build the recognition model. Identifying alien appliances is an interesting problem as it gives flexibility to the consumer to introduce new household appliances without providing any additional information to the service provider. In other words, the model should rely upon only the semantic information of native (seen) appliances to identify the alien ones. We plan to develop effective strategies for extracting semantic information, thereby enhancing the capability of our recognition model in FedAR+.



{\em Malicious client:} A client with wrong intention may try to attack the model by altering its weight parameters when transmitted from the client to server. Such malicious client can affect the performance of the global model in FedAR+. We plan to detect the malicious clients who send incorrect parameters by either adding random-noise or backdoor patterns in the dataset. Our idea is to exploit the history of each client' gradients with an appropriate similarity measures (e.g., cosine distance) to distinguish malicious clients from the normal ones and exclude their parameters from the aggregation.  

  

\bibliographystyle{ACM-Reference-Format}
\bibliography{main}

\end{document}